\documentclass{article} 
\usepackage[dvipsnames]{xcolor}
\usepackage{iclr2024_conference,times}

\usepackage{amsmath,amsfonts,bm}

\def\eqref#1{equation~\ref{#1}}

\def\1{\bm{1}}

\DeclareMathAlphabet{\mathsfit}{\encodingdefault}{\sfdefault}{m}{sl}
\SetMathAlphabet{\mathsfit}{bold}{\encodingdefault}{\sfdefault}{bx}{n}

\usepackage{hyperref}
\usepackage{url}

\usepackage[frozencache,cachedir=.]{minted}
\newminted{python}{frame=lines,framerule=2pt}
\usepackage{graphicx}
\usepackage{booktabs}
\usepackage{algorithm}
\usepackage[noend]{algpseudocode}
\newcommand{\envpool}{EnvPool}

\newcommand{\hflogo}{{\includegraphics[scale=0.13]{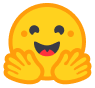}}}

\colorlet{LightAquamarine}{Aquamarine!40!White}
\colorlet{LightSalmon}{Salmon!40!White}
\colorlet{LightGoldenrod}{Goldenrod!40!White}
\colorlet{LightOrchid}{Orchid!40!White}
\definecolor{myblue}{HTML}{7F94C4}
\definecolor{myorange}{HTML}{DCA67E}
\colorlet{lightmyblue}{myblue!60!White}
\colorlet{lightmyorange}{myorange!60!White}
\definecolor{myred}{HTML}{C97F7E}
\definecolor{mybrown}{HTML}{AB9B88}
\colorlet{lightmyred}{myred!60!White}
\colorlet{lightmybrown}{mybrown!60!White}
\newcommand{\mysmall}{\fontsize{8pt}{9.6pt}\selectfont}

\title{Cleanba: A Reproducible and Efficient Distributed Reinforcement Learning Platform}

\author{Shengyi Huang\textsuperscript{\ddag}\hflogo \quad 
  Jiayi Weng\thanks{Currently at OpenAI.} \quad Rujikorn Charakorn\textsuperscript{$\sharp$} \quad Min Lin\textsuperscript{$\triangle$}  \\
   \textbf{Zhongwen Xu\textsuperscript{$\diamondsuit$} \quad Santiago Onta\~{n}\'{o}n\textsuperscript{\ddag}\textsuperscript{$\mathsection$}} \\
   \\
   \textsuperscript{\ddag}Drexel University \quad \hflogo Hugging Face \quad \textsuperscript{$\mathsection$}Google \quad \textsuperscript{$\sharp$}VISTEC \quad 
   \textsuperscript{$\triangle$}Sea AI Lab \quad
   \textsuperscript{$\diamondsuit$}Tencent AI Lab
   \\
  \texttt{costa.huang@outlook.com}
}

\iclrfinalcopy 
\begin{document}

\maketitle


\begin{abstract}
Distributed Deep Reinforcement Learning (DRL) aims to leverage more computational resources to train autonomous agents with less training time. Despite recent progress in the field, reproducibility issues have not been sufficiently explored. This paper first shows that the typical actor-learner framework can have reproducibility issues even if hyperparameters are controlled. We then introduce Cleanba, a new open-source platform for distributed DRL that proposes a highly reproducible architecture. Cleanba implements highly optimized distributed variants of PPO~\citep{schulman2017proximal} and IMPALA~\citep{espeholt2018impala}. Our Atari experiments show that these variants can obtain equivalent or higher scores than strong IMPALA baselines in \texttt{moolib} and \texttt{torchbeast} and PPO baseline in CleanRL. However, Cleanba variants present 1) shorter training time and 2) more reproducible learning curves in different hardware settings. Cleanba's source code is available at \url{https://github.com/vwxyzjn/cleanba}
\end{abstract}

\section{Introduction}
Deep Reinforcement Learning (DRL) is a technique to train autonomous agents to perform tasks. In recent years, it has demonstrated remarkable success across various domains, including video games~\citep{mnih2015human}, robotics control~\citep{schulman2017proximal}, chip design~\citep{mirhoseini2021graph}, and large language model tuning~\citep{ouyang2022training}. Distributed DRL~\citep{espeholt2018impala,Espeholt2020SEED} has also become a fast-growing field that leverages more computing resources to train agents. Despite recent progress, reproducibility issues in distributed DRL have not been sufficiently explored. This paper introduces Cleanba, a new platform for distributed DRL that addresses reproducibility issues under different hardware settings. 

Reproducibility in DRL is a challenging issue. Not only are DRL algorithms brittle to hyperparameters and neural network architectures~\citep{henderson2018deep}, implementation details are often crucial for successfully applying DRL but frequently omitted from publications~\citep{Engstrom2020Implementation,andrychowicz2021what,shengyi2022the37implementation}. Reproducibility issues in distributed DRL are under-studied and arguably even more challenging. In particular, most high-profile distributed DRL works, such as Apex-DQN~\citep{horgan2018distributed}, IMPALA~\citep{espeholt2018impala}, R2D2~\citep{kapturowski2019recurrent}, and Podracer Sebulba~\citep{hessel2021podracer} are not (fully) open-source. Furthermore, earlier work pointed out that more actor threads not only improve training speed but cause reproducibility issues -- different hardware settings could impact the data efficiency in a non-linear fashion~\citep{mnih2016asynchronous}. 

In this paper, we present a more principled approach to distributed DRL, in which different hardware settings could make training speed slower or faster but do not impact data efficiency, thus making scaling results more reproducible and predictable. We first analyze the typical actor-learner architecture in IMPALA~\citep{espeholt2018impala} and show that its parallelism paradigm could introduce reproducibility issues due to the concurrent scheduling of different actor threads. We then propose a more reproducible distributed architecture by better aligning the parallelized actor and learner's computations. Based on this architecture, we introduce our Cleanba (meaning \textbf{Clean}RL-style~\citep{huang2022cleanrl} Podracer Sebul\textbf{ba}) distributed DRL platform, which aims to be an easy-to-understand distributed DRL infrastructure like CleanRL, but also be scalable as Podracer Sebulba. Cleanba implements a distributed variant of PPO~\citep{schulman2017proximal} and IMPALA~\citep{espeholt2018impala} with JAX~\citep{bradbury2018jax} and EnvPool~\citep{weng2022envpool}. Next, we evaluate Cleanba's variants against strong IMPALA baselines in \texttt{moolib}~\citep{moolib2022} and torchbeast~\citep{kuttler2019torchbeast} and PPO baseline in CleanRL~\citep{huang2022cleanrl} on 57 Atari games~\citep{bellemare2013arcade}. Here are the key results of Cleanba:
\begin{enumerate}
    \item \textbf{Strong performance}: Cleanba's IMPALA and PPO achieve about 165\% median human normalized score (HNS) in Atari with sticky actions, matching \texttt{monobeast} IMPALA's 165\% median HNS and outperforming \texttt{moolib} IMPALA's 140\% median HNS.
    \item \textbf{Short training time}: Under the 1 GPU 10 CPU setting, Cleanba's IMPALA is \textbf{6.8x faster} than \texttt{monobeast}'s IMPALA and \textbf{1.2x faster} than \texttt{moolib}'s IMPALA. Under a max specification setting, Cleanba's IMPALA (8 GPU and 40 CPU) is \textbf{5x faster} than \texttt{monobeast}'s IMPALA (1 GPU and 80 CPU) and \textbf{2x faster} than \texttt{moolib}'s IMPALA (8 GPU and 80 CPU).
    \item \textbf{Highly reproducible}: Cleanba shows predictable and reproducible learning curves across 1 and 8 GPU settings given the same set of hyperparameters, whereas \texttt{moolib}'s learning curves can be considerably different, even if hyperparameters are controlled to be the same.
\end{enumerate}
To facilitate more transparency and reproducibility, we have made available our source code at \url{https://github.com/vwxyzjn/cleanba}.

\section{Background}
\textbf{Distributed DRL Systems } Utilizing more computational power has been an attractive topic for researchers. Earlier DRL methods like DQN~\citep{mnih2015human} were synchronous and typically used a single simulation environment, which made them slow and inefficient in using hardware resources. A3C~\citep{mnih2016asynchronous} spawns multiple actor threads; each interacts with its own copy of the environment and asynchronously accumulates gradient.  To make distributed DRL more scalable, IMPALA decouples the actors and the learners~\citep{espeholt2018impala,Espeholt2020SEED}. The actors produce training data asynchronously, while the learners produce new agent parameters, which are transferred asynchronously to the actor. Actor-learner systems can achieve higher throughput and shorter training wall time than A3C. Additional distributed actor-learner systems include GA3C~\citep{babaeizadeh2017reinforcement}, IMPALA~\citep{espeholt2018impala}, Apex-DQN~\citep{horgan2018distributed}, R2D2~\citep{kapturowski2019recurrent}, and Podracer Sebulba~\citep{hessel2021podracer}.

\textbf{Reproducibility Issues with Different Hardware Settings }
Empirical evidence suggests that increasing the number of actor threads can enhance the training speed in distributed DRL (\citet[Fig. 4]{mnih2016asynchronous}). However, this augmentation is not without its complications. It also impacts data efficiency and final Atari scores (\citet[Fig. 3]{mnih2016asynchronous}), and these effects could manifest in a non-linear manner. While the authors found the side effects of value-based asynchronous methods to be positive and improve data efficiency, the side effects of contemporary distributed DRL systems, such as IMPALA, Apex-DQN, and R2D2, across various hardware configurations, have not been sufficiently explored.

\textbf{Open-source Distributed DRL 
Infrastructure}
While many distributed DRL algorithms are not open-source, there have been many notable distributed DRL replications in the open-source software (OSS) community. These efforts include SEED RL~\citep{Espeholt2020SEED}, \texttt{rlplyt}~\citep{stooke2018accelerated}, Decentralized Distributed PPO~\citep{Wijmans2020DD-PPO}, Sample Factory~\citep{petrenko2020sf}, HTS-RL~\citep{liu2020high}, \texttt{torchbeast}~\citep{kuttler2019torchbeast}, and \texttt{moolib}~\citep{moolib2022}. Many of them have shown high throughput and good empirical performance in select domains. Nevertheless, most of them either do not have evaluations on 57 Atari games or have various hardware restrictions, leading to reproducibility concerns. \texttt{moolib} is the only OSS infrastructure that has both evaluations on 57 Atari games in the standard 200M frames setting and can scale beyond a single GPU setting\footnote{While SEED RL also has evaluations on 57 Atari games and scale beyond 1 GPU, SEED RL trained the agents for 40 billion frames 40 hours per game. }.

\section{Reproducibility Issues in IMPALA}
\label{sec:impala-reproducibility-issues} 
This section shows that IMPALA~\citep{espeholt2018impala} has non-determinism by nature, which arises from the concurrent scheduling of different actor threads. This non-determinism could further cause subtle reproducibility issues.

\begin{figure}[t]
\begin{minipage}[t]{\linewidth}
    \footnotesize
    \begin{minipage}[t]{0.50\linewidth}
        \paragraph{IMPALA Actor-Learner Architecture}
        \label{lst:branching_code}
\begin{minted}[frame=single,framesep=10pt,highlightlines={2-3,5-9,12,13,15,16-17},highlightcolor=Lavender,linenos,fontsize=\mysmall]{python}
batch_size = 32
agent = Agent()
data_Q = queue()

def actor():
  while True:
    data = rollout(agent.param, 1)


    data_Q.put(data)
def learner():
  for _ in range(1, ITER):
    data = data_Q.get_many(batch_size)
    agent.learn(data)
    broadcast_to_actors(agent.param)
for _ in range(num_actors):
  thread(actor).start()
thread(learner).start()
\end{minted}
    \end{minipage}
    \hfill
    \begin{minipage}[t]{0.50\linewidth}
        \paragraph{Cleanba's architecture}
        \label{lst:branchless_code}
\begin{minted}[frame=single,framesep=10pt,highlightlines={2-3,5-9,12,13,15,16-17},highlightcolor=YellowGreen,fontsize=\mysmall]{python}
batch_size = 32
agent = Agent()
data_Q = queue(max_size=1)
param_Q = queue(max_size=1)
def actor():
  for i in range(1, ITER):
    if i != 2:
      params = param_Q.get()
    data = rollout(params, batch_size)
    data_Q.put(data)
def learner():
  for _ in range(1, ITER):
    data = data_Q.get()
    agent.learn(data)
    param_Q.put(agent.param)
param_Q.put(agent.param)
thread(actor).start()
thread(learner).start()
\end{minted}
    \end{minipage}
\end{minipage}
\caption{The pseudocode for IMPALA architecture (left) and Cleanba's architecture (right). Colors are used to highlight the code differences between the two architectures. The \texttt{rollout(params, num\_envs)} function collects rollout data on \texttt{num\_envs} independent environments for  \texttt{num\_steps} steps.}
\label{fig:impala-pseudocode}
\end{figure}

A natural question arises: \emph{what happens when the learner produces a new policy while the actor is in the middle of producing a trajectory?} It turns out multiple policy versions could contribute to the actor's rollout data in line 7 of the IMPALA architecture Figure~\ref{fig:impala-pseudocode}. Typically, the faster the policy updates, the more frequently the policies are transferred. However, this impacts the rollout data construction in a non-trivial way. From a reproducibility point of view, it is important to realize the frequency at which the policies are updated is a source of non-determinism. 

However, non-determinism can be desirable in parallel programming because they make programs faster without making outputs significantly different. For example, some of NVIDIA's CuDNN operations are inherently non-determinisitic\footnote{\url{https://docs.nvidia.com/deeplearning/cudnn/developer-guide/index.html\#reproducibility}}. What is more important is to investigate if this non-determinism could cause reproducibility issues in terms of learning curves.
%
\begin{figure}[t]
    \centering
    \includegraphics[width=0.90\linewidth]{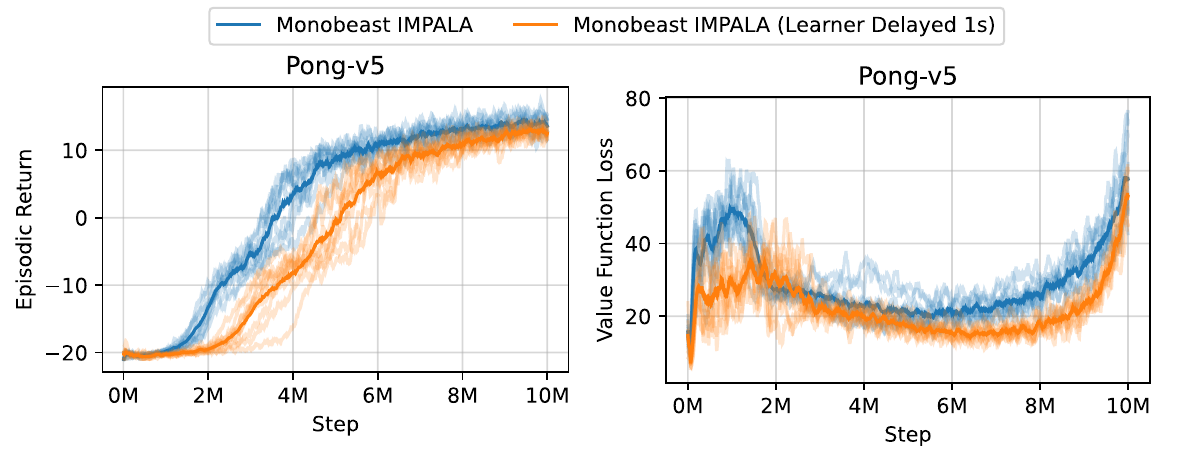}
    \caption{\textbf{IMPALA's reproducibility issue under different ``speed'' settings} — The y-axes show the episodic return and value function loss of two sets of  \texttt{monobeast}  experiments that use the \emph{exact same hyperparameters}, but the orange set of experiments has its learner update manually delayed for 1 second to simulate slower learner updates. Note the learning curves across 10 random seeds are non-trivially different, implicating hyperparameters in IMPALA alone cannot always ensure good reproducibility.}
    \label{fig:torchbeast_reproducibility}
\end{figure}
To this end, we manufacture a specific experiment that magnifies this non-determinism in \texttt{monobeast}'s IMPALA. For the control group, we 
\begin{enumerate}
    \item decreased the number of trajectories in the batch from 32 to 8 to reduce training time, thus making the actor's policy updates more frequent;
    \item used 80 actor threads and increased \texttt{monobeast}'s default unroll length from 20 to 240 to increase the chance of observing the actor's policy updates in the middle of a trajectory.
\end{enumerate}
For the experimental group, we used the above setting but \emph{manually slowed down the policy broadcasting} by sleeping the learner for 1 second after the policy updates in order to simulate a case where the learner is significantly slower (such as when running the learner on CPU).

We found that in the control group, the actors, on average, changed their policy versions \textbf{12-13 times} in the middle of the 240-length trajectory. In the experimental group, because of the manual slowdown in broadcasting the learner's policy, the actors, on average, changed the policy \textbf{one time}. We note that the results vary on different hardware settings as well. For example, the control group changed their policy versions, on average, eight times when using 40 actor threads. We noted that in \texttt{moolib}, the actor's policy could also change mid-rollout. See Appendix~\ref{sec:torchbeast_logs}.

Figure~\ref{fig:torchbeast_reproducibility} demonstrates the empirical effect of the experiments. Note that the learning and loss curves looked notably different across ten random seeds, even though the control and experimental group have the \emph{exact same hyperparameters}. This experiment shows that IMPALA algorithmically could be susceptible to reproducibility issues across different hardware settings. While Figure~\ref{fig:torchbeast_reproducibility} only shows the experimental results on one environment, the primary purpose of it is to show that this issue exists and is barely predictable. Furthermore, this type of issue can be much more subtle and difficult to diagnose at a much larger scale, so it is important that we investigate them.

\section{Towards Reproducible Distributed DRL}
\label{sec:cleanba-arch}
Despite these reproducibility issues, the actor-learner architecture is useful because it allows us to parallelize the computations of the actors and learners. In this work, we address the reproducibility issues mentioned above by 1) decoupling hyperparameters and hardware settings and 2) proposing a synchronization mechanism that makes distributed DRL reproducible.

\subsection{Decoupling hyperparameters and hardware settings} As mentioned in the previous section, different numbers of actor threads could make policy updates more or less frequent in the middle of a trajectory generation.  This is unpredictable and need not be the case. A different number of actors also creates a different number of simulation environments and thus should be recognized as a hyperparameter setting. 

To make a more clarified setting, we advocate decoupling the number of actor threads into two separate hyperparameters: 1) the number of environments, and 2) the number of CPUs. In this case,  we can use a different number of CPUs to simulate a given number of environments. This decoupled interface is readily provided by EnvPool~\citep{weng2022envpool}, which we use in our proposed architecture.

\subsection{Deterministic Rollout Data Composition}
To address the non-determinism in rollout data composition, we propose our \emph{Cleanba's architecture}, which retains the benefit of parallelizing actor-learner computations but can produce deterministic rollout data composition. At its core, Cleanba's architecture is a simple mechanism for synchronizing the actor and learner, ensuring the learner performs gradient updates with rollout data of \textbf{second latest policy}.





Let us use the notation $\pi_i \rightarrow \mathcal{D}_{\pi_i}$ to denote that policy of version $i$ is used to obtain rollout data $\mathcal{D}_{\pi_i}$; $\pi_{i} \xrightarrow[\mathcal{D}_{\pi_i}]{} \pi_{i+1}$ denotes policy of version $i$ is trained with rollout data $\mathcal{D}_{\pi_i}$ to obtain a new policy $\pi_{i+1}$. Figure~\ref{fig:impala-pseudocode} is the pseudocode of the architecture and Table~\ref{tab:rollout-data-composition} illustrates how policies get updated.
Under the Synchronous Architecture, the actor and learner's computations are sequential: it first perform rollout \colorbox{LightAquamarine}{$\pi_1 \rightarrow \mathcal{D}_{\pi_1}$}, during which the learner stays idle. Given the rollout data, the learner then performs gradient updates \colorbox{LightSalmon}{$\pi_{1} \xrightarrow[\mathcal{D}_{\pi_1}]{} \pi_2$}, during which the actor stays idle. More generally, the learner always learns from the rollout data of the latest policy  $\pi_{i} \xrightarrow[\mathcal{D}_{\pi_{i}}]{}  \pi_{i+1}$.

To parallelize actor and learner's computation, Cleanba's architecture needs to necessarily introduce stale data like IMPALA~\citep{espeholt2018impala}. In the second iteration of Cleanba's architecture in Figure~\ref{fig:impala-pseudocode}, we skip the \texttt{param\_Q.get()} call, so \colorbox{LightAquamarine}{$\pi_1 \rightarrow \mathcal{D}_{\pi_1}$} happens concurrently with \colorbox{LightSalmon}{$\pi_{1} \xrightarrow[\mathcal{D}_{\pi_1}]{} \pi_2$}. Because \texttt{Queue.get} is blocking when the queue is empty and \texttt{Queue.put} is blocking when the queue is full (we set the maximum size to be 1), we make sure the actor process does not perform more rollouts and learner process does not perform more gradient updates. Starting iteration $i>3$, the learner then learns from the rollout data of the second latest policy $\pi_{i} \xrightarrow[\mathcal{D}_{\pi_{i-1}}]{}  \pi_{i+1}$. As a result, Cleanba's architecture can parallelize the actor and learner's computation at the cost of stale data.


\begin{table}[t]
\centering
\caption{The Synchronous and Cleanba's architecture. 
Under the Synchronous architecture, the actor and learner's computations are sequential and \emph{not} parallelizable -- the learner always learns from the rollout data of the latest policy $\pi_{i} \xrightarrow[\mathcal{D}_{\pi_{i}}]{}  \pi_{i+1}$ (e.g., \colorbox{LightOrchid}{$\pi_{2} \xrightarrow[\mathcal{D}_{\pi_2}]{}  \pi_3$}). Under Cleanba's architecture, we can parallelize the actor and learner's computation at the cost of introducing stale data -- starting from iteration 3 the learner always learns from the rollout data obtained from the second latest policy $\pi_{i} \xrightarrow[\mathcal{D}_{\pi_{i-1}}]{}  \pi_{i+1}$ (e.g., \colorbox{Orchid}{$\pi_{2} \xrightarrow[\mathcal{D}_{\pi_1}]{}  \pi_3$})}
\label{tab:rollout-data-composition}
\begin{small}
\begin{tabular}{@{}lccc@{}}
\toprule
Iteration & 1 & 2 & 3  \\ 
\midrule
Synchronous Arch. & \colorbox{LightAquamarine}{$\pi_1 \rightarrow \mathcal{D}_{\pi_1}$}\colorbox{LightSalmon}{$\pi_{1} \xrightarrow[\mathcal{D}_{\pi_1}]{} \pi_2$} & \colorbox{LightGoldenrod}{$\pi_2 \rightarrow \mathcal{D}_{\pi_2}$}\colorbox{LightOrchid}{$\pi_{2} \xrightarrow[\mathcal{D}_{\pi_2}]{}  \pi_3$} &$\pi_3 \rightarrow \mathcal{D}_{\pi_3}$ $\pi_{3} \xrightarrow[\mathcal{D}_{\pi_3}]{}  \pi_4$ \\
\midrule
Cleanba's Arch., Actor & \colorbox{LightAquamarine}{$\pi_1 \rightarrow \mathcal{D}_{\pi_1}$} & \colorbox{LightAquamarine}{$\pi_1 \rightarrow \mathcal{D}_{\pi_1}$} & \colorbox{LightGoldenrod}{$\pi_2 \rightarrow \mathcal{D}_{\pi_2}$}  \\ 
Cleanba's Arch., Learner & & \colorbox{LightSalmon}{$\pi_{1} \xrightarrow[\mathcal{D}_{\pi_1}]{}  \pi_2$} & \colorbox{Orchid}{$\pi_{2} \xrightarrow[\mathcal{D}_{\pi_1}]{}  \pi_3$}  \\
\bottomrule
\end{tabular}
\end{small}
\end{table}

Cleanba's architecture above has several benefits. First, it is easy to reason and reproduce. As highlighted in Table~\ref{tab:rollout-data-composition}, we can ascertain the specific policy used for collecting the rollout data, 
so if we had delayed learner updates like in Section~\ref{sec:impala-reproducibility-issues} for iteration $i$, iteration $i+1$ would not start until the previous iteration is finished, therefore circumventing IMPALA's reproducibility issue.
This knowledge about which policy generates the rollout data enhances the transparency and reproducibility of distributed RL and can help us scale up while maintaining good reproducibility principles. Second, Cleanba's architecture is easy to debug for throughput. For diagnosing throughput, we can evaluate the time taken for \texttt{rollout\_Q.get()} and \texttt{param\_Q.get()}. If, on average, \texttt{rollout\_Q.get()} consumes less time than \texttt{param\_Q.get()}, it becomes evident that learning is the bottleneck, and vice versa.

\begin{figure}[!t]
    \centering
    \includegraphics[width=0.90\linewidth]{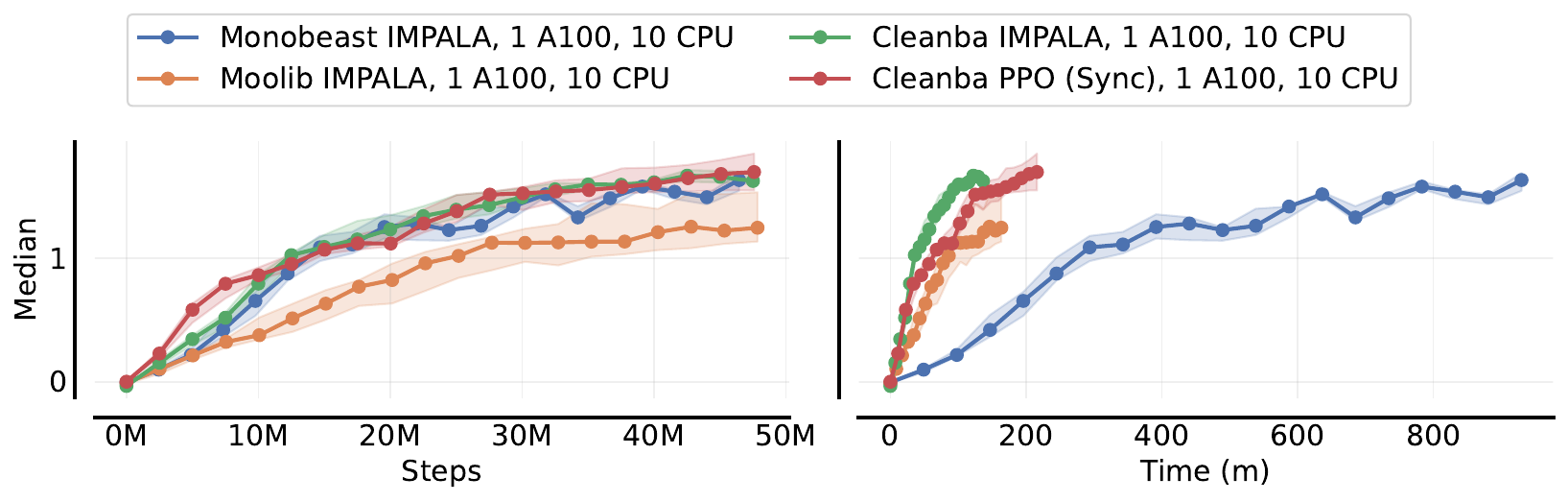}
    \includegraphics[width=0.99\linewidth]{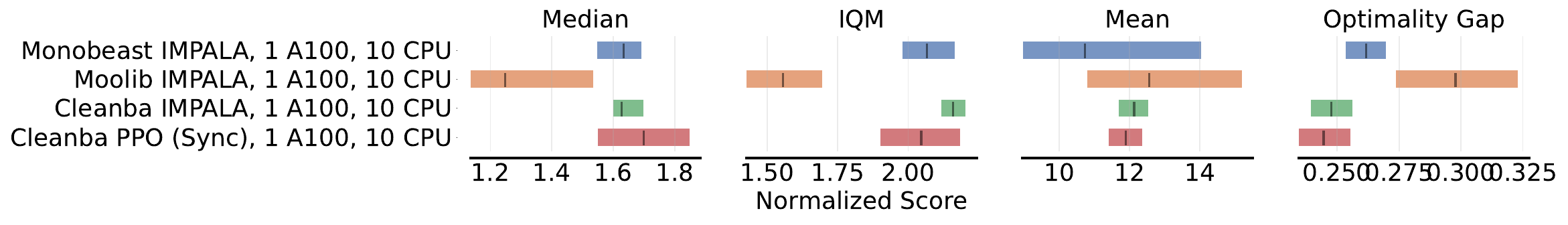}
    \caption{Base experiments. Top figure: the median human-normalized scores of Cleanba variants compared with \texttt{moolib} and \texttt{monobeast}. Bottom figure: the  aggregate human normalized score metrics with 95\% stratified bootstrap CIs. Higher is better for Median, IQM, and Mean; lower is better for Optimality Gap.}
    \label{fig:10cpuexp}
\end{figure}
\begin{figure}[!t]
    \centering
    \includegraphics[width=0.90\linewidth]{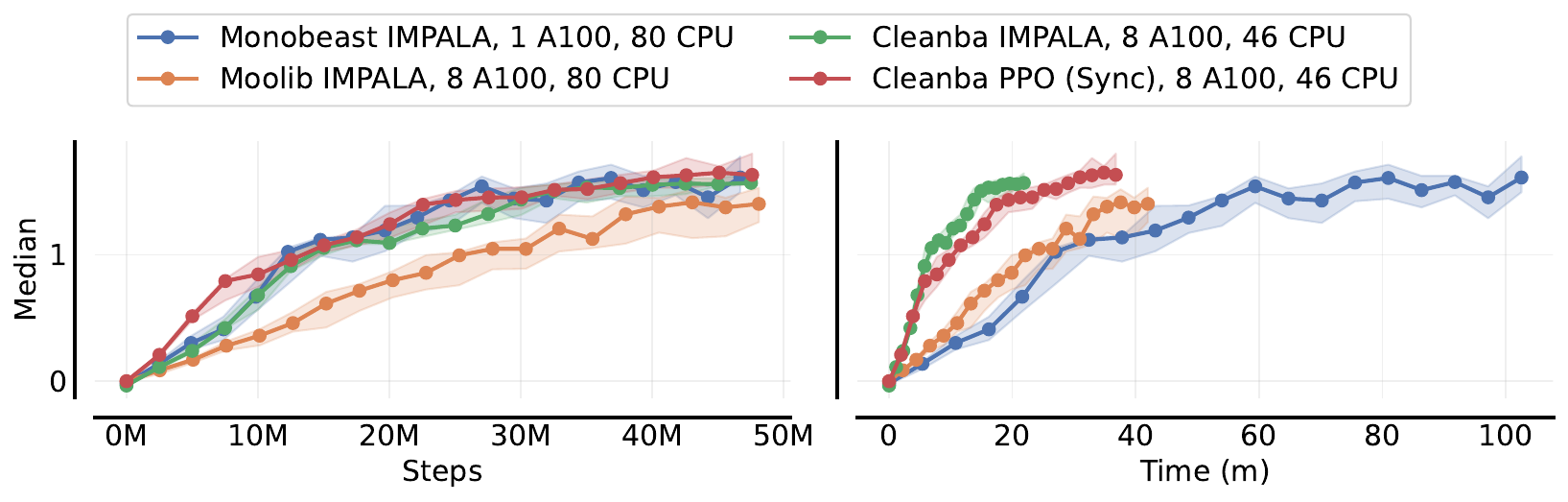}
    \includegraphics[width=0.99\linewidth]{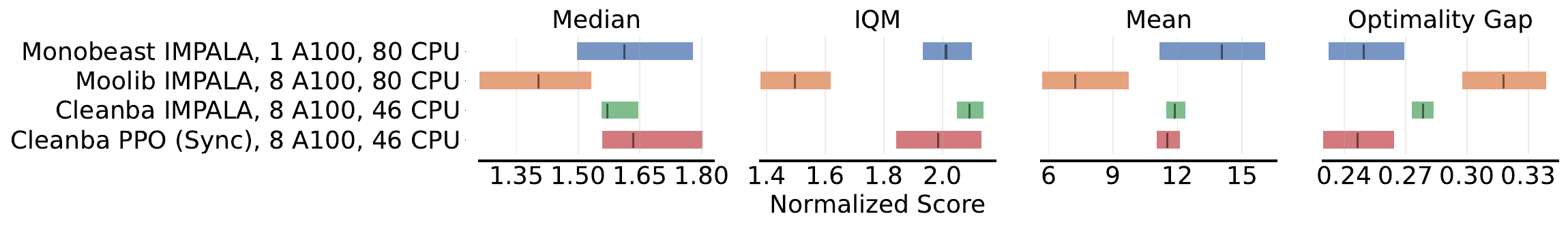}
    \caption{Workstation experiments. Top figure: the median human-normalized scores of Cleanba variants compared with \texttt{moolib}. Bottom figure: the aggregate human normalized score metrics with 95\% stratified bootstrap CIs.}
    \label{fig:specoutexp}
\end{figure}

Based on Cleanba's architecture, this work introduces Cleanba as a reproducible distributed DRL platform. Cleanba is inspired by CleanRL~\citep{huang2022cleanrl} and DeepMind's Sebulba Podracer architecture~\citep{hessel2021podracer}. Its implementation uses JAX~\citep{bradbury2018jax} and EnvPool~\citep{weng2022envpool}, both of which are designed to be efficient. 
To improve the learner's throughput, we allow the use of multiple learner devices via \texttt{pmap}. To improve the system's scalability, we enable running multiple processes on a single node or multiple nodes via \texttt{jax.distibuted}.

\section{Experiments}
\label{sec:experiments-atari}
We perform experiments on Atari games~\citep{bellemare2013arcade}. All experiments used $84\times84$ images with greyscale, an action repeat of 4, 4 stacked frames, and a maximum of 108,000 frames per episode. We followed the recommended Atari evaluation protocol by \citet{machado2018revisiting}, which used sticky action with a probability of 25\%, no loss of life signal, and the full action space. To make a more direct and fair comparison, we used the same AWS p4d.24xlarge instances\footnote{For some experiments, we used p4de.24xlarge instances but only GPU memory is different, which does not affect training speed. } and the same Atari environment simulation setups via \envpool~and compared only the following codebase settings:
\begin{enumerate}
    \item \textbf{Monobeast IMPALA}: the reference IMPALA implementations in \texttt{monobeast}\footnote{We wanted to test out IMPALA's official source code released in \texttt{deepmind/scalable\_agent}, but it was built with \texttt{tensorflow 1.x} which does not support the A100 GPU tested in this paper.};
    \item \textbf{Moolib IMPALA}: the reference IMPALA implementations in \texttt{Moolib};
    \item \textbf{CleanRL PPO (Sync)}: the reference PPO implementations in CleanRL\citep{huang2022cleanrl};
    \item \textbf{Cleanba PPO} and \textbf{Cleanba IMPALA}: our PPO and IMPALA implementation under the Cleanba Architecture;
    \item \textbf{Cleanba PPO (Sync)} and \textbf{Cleanba IMPALA (Sync)} our PPO and IMPALA implementation under the Synchronous Architecture (Table~\ref{tab:rollout-data-composition}), which can be configured by commenting out line 7 of the Cleanba's architecture in Figure~\ref{fig:impala-pseudocode}.
\end{enumerate}
Within the p4d.24xlarge instance, we also compared two hardware settings:
\begin{enumerate}
    \item \textbf{Base experiments} uses 10 CPU and 1 A100 setting as a base comparison;
    \item \textbf{Workstation experiments} uses 46 CPU and 8 A100s for Cleanba experiments, 80 CPU and 8 A100s for \texttt{moolib} experiments\footnote{We used more CPUs for \texttt{moolib} experiments because 10 CPU per GPU seems to be the default scaling parameter for \texttt{moolib}. Also, for the \texttt{moolib} experiment, we conducted two sets of 3 random seeds. We reported the results with higher IQM and lower median. See Appendix~\ref{appendix:moolib_twosets}.}, and 80 CPU and 1 A100 for \texttt{monobeast} experiments.
\end{enumerate}
\begin{figure}[t]
    \centering
    \includegraphics[width=0.95\textwidth]{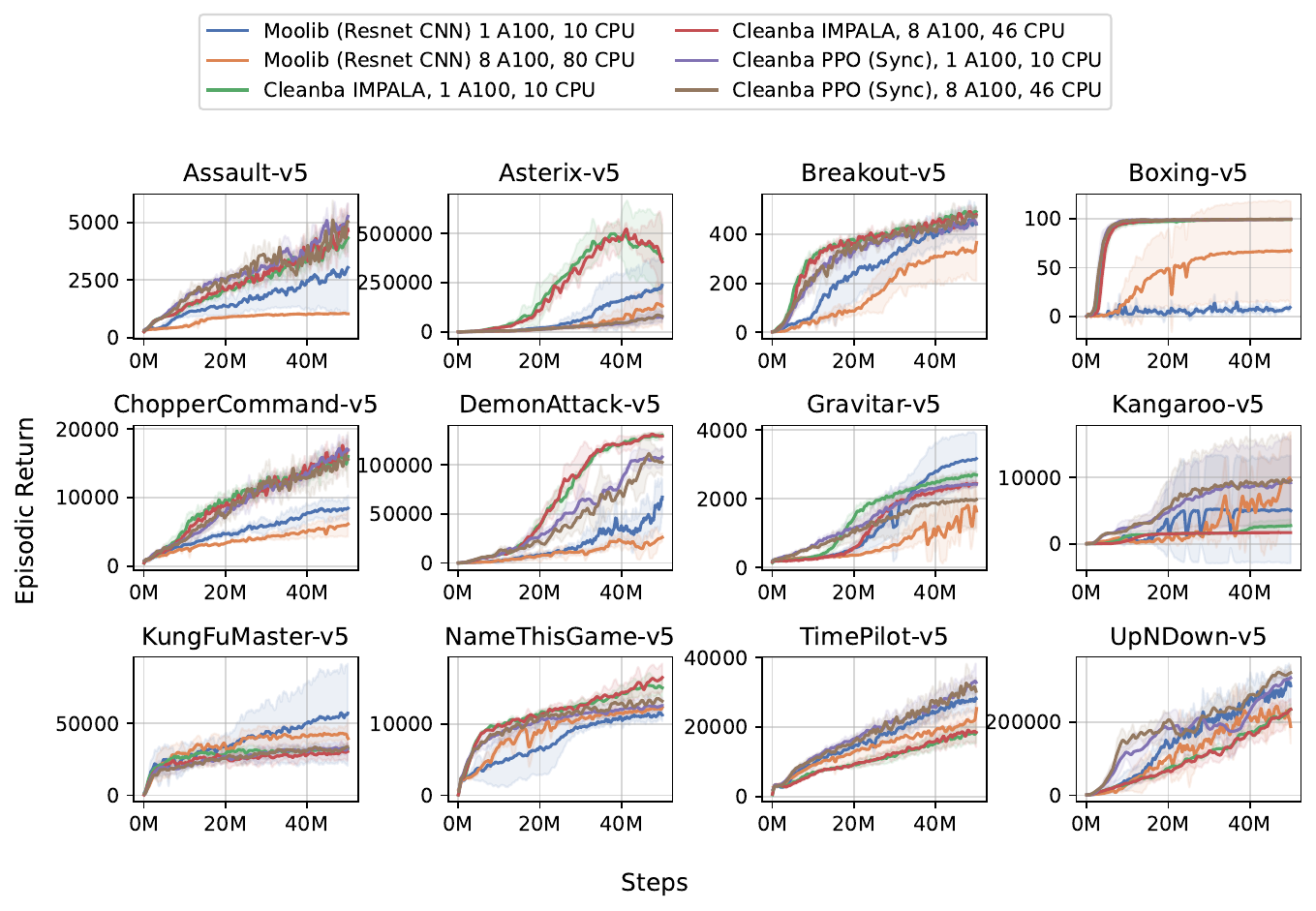}
    \includegraphics[width=0.99\linewidth]{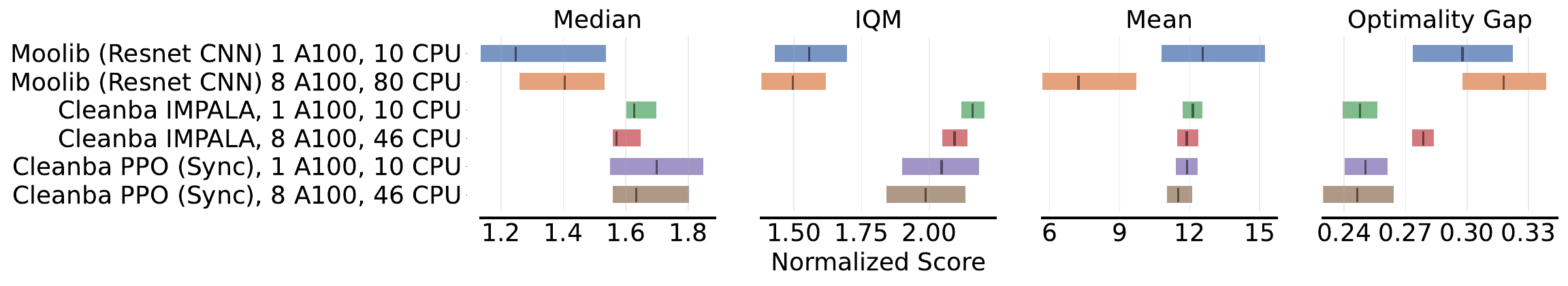}
    \caption{Reproducible learning curves -- the Cleanba variants show more predictable learning curves in different hardware settings. In comparison, \texttt{moolib}'s IMPALA's learning curves under the \colorbox{lightmyblue}{1 A100, 10 CPU setting (blue curve)} and \colorbox{lightmyorange}{8 A100, 80 CPU setting (orange curve)} are meaningfully different, even if they use the same hyperparameters.}
    \label{fig:smoothness}
\end{figure}

Throughout all experiments, the agents used IMPALA's Resnet architecture~\citep{espeholt2018impala}, ran for 200M frames with three random seeds. The hyperparameters and the learning curves can be found in Appendix~\ref{appendix:detailed-setting}. We evaluate the experiment results based on median HNS learning curves, interquartile mean (IQM) learning curves, and 95\% stratified bootstrap confidence intervals for the mean, median, IQM, and optimality gap (the amount by which the algorithm fails to meet a minimum normalized score of 1)~\citep{agarwal2021deep}.

\subsection{Comparison with \texttt{moolib} and \texttt{monobeast}'s IMPALA}
Under the base experiments (Figure~\ref{fig:10cpuexp}), Cleanba’s IMPALA obtains a similar level of median HNS as \texttt{monobeast}'s IMPALA and a higher level of median HNS as \texttt{moolib}'s IMPALA. However, Cleanba's IMPALA is  \textbf{6.8x faster} than \texttt{monobeast}'s IMPALA, mostly because Cleanba actors run on GPUs, whereas \texttt{monobeast}'s actors run on CPUs. Also, Cleanba's IMPALA is \textbf{1.2x faster} than \texttt{moolib}'s IMPALA, but the speedup difference is challenging to explain due to multiple confounding factors -- Cleanba's variants benefit from JAX's just-in-time compilation, whereas \texttt{moolib} benefits from asynchronous operations (e.g., on gradient computation and environment steps). Cleanba's PPO (Sync) also obtains a high median HNS but takes longer training time, likely due to the longer training step time spent on reusing rollout data 4 times.

Under the workstation experiments (Figure~\ref{fig:specoutexp}), Cleanba’s PPO (Sync) and IMPALA obtain a similar level of median HNS as \texttt{monobeast}'s IMPALA and a higher level of median HNS as \texttt{moolib}'s IMPALA. However, Cleanba's PPO (Sync)  and IMPALA are both faster than \texttt{monobeast}'s and \texttt{moolib} IMPALA. Most prominently, Cleanba's IMPALA is \textbf{5x faster} than \texttt{monobeast}'s IMPALA and \textbf{2x faster} than \texttt{moolib}'s IMPALA.

Additionally, we examine the individual learning curves in Figure~\ref{fig:smoothness} and found that Cleanba's variants also produce more consistent learning curves. In comparison, in two hardware settings, \texttt{moolib}'s learning curves can be much more unpredictable.

\subsection{Discussion about \texttt{monobeast}'s IMPALA}
Note that the \texttt{monobeast} experiments are interesting in several ways. First, it produces a higher median HNS than \texttt{moolib}'s IMPALA, which is the opposite of what was shown in \citet{moolib2022}. This is probably because \citet{moolib2022} used ``comparable environment settings'' instead of the same environment settings used in our experiments. Interestingly, we found different Atari wrapper implementations can have a non-trivial impact on the agent's performance (Appendix~\ref{appendix:wrapper-matter}); for this reason, we use the same Atari wrapper implementation in the experiments presented in this section. Second, the \texttt{monobeast} experiments appear robust in two different hardware settings in practice, despite the reproducibility issues we showed in Section~\ref{sec:impala-reproducibility-issues}. While \texttt{monobeast} obtained high scores, it is significantly slower in the 1 A100 and 10 CPU settings due to poor GPU utilization. Its codebase also does not support multi-GPU settings and should scale less efficiently with larger networks because actor threads only run on CPUs when compared to \texttt{moolib} and Cleanba's variants.

\begin{figure}[t]
\centering
    \includegraphics[width=0.90\linewidth]{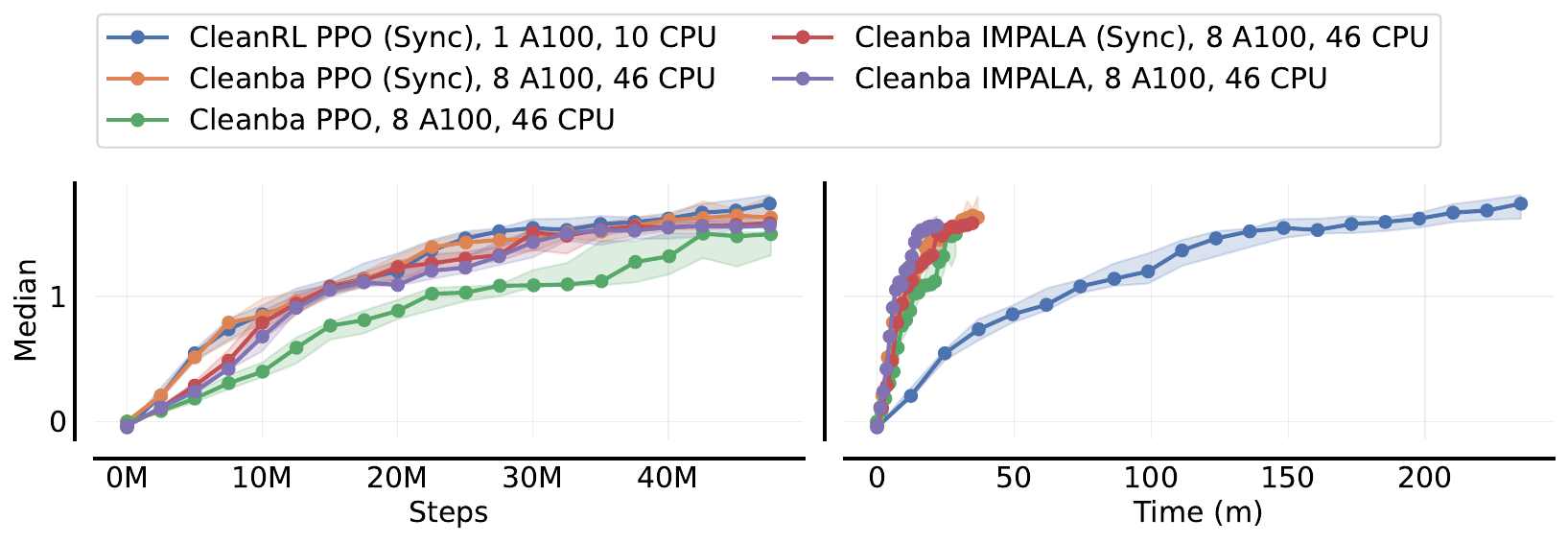}
    \includegraphics[width=0.99\linewidth]{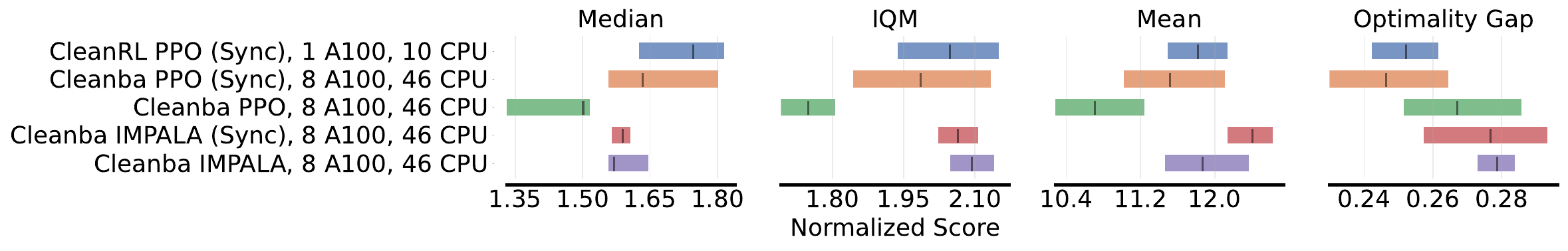}
    \caption{Comparing Cleanba's variants using Cleanba and Synchronous architecture. For PPO, \colorbox{lightmyorange}{Cleanba's Architecture (orange curve)} runs faster but has lower data efficiency than \colorbox{lightmyblue}{Synchronous architecture (blue curve)}. For IMPALA, there is no discernible difference between  \colorbox{lightmyred}{Synchronous Architecture (red curve)} and \colorbox{lightmybrown}{Cleanba's Architecture (brown curve)}. This means Cleanba's IMPALA can benefit from the speed-up of parallelizing actor-learner computation without paying a price for data efficiency under our hyperparameter settings, unlike Cleanba's PPO.}
    \label{fig:compared_with_cleanrl}
\end{figure}

\subsection{Synchronous Architecture vs Cleanba Architecture}
Figure~\ref{fig:compared_with_cleanrl} compares the PPO and IMPALA variants between Synchronous and Cleanba architecture and CleanRL's PPO, which uses the Synchronous architecture by design. We found using Cleanba architecture actually hurts Cleanba PPO's data efficiency. This is an interesting trade-off because the speed benefit of parallelizing actor and learner processes in Cleanba PPO is offset by the lower data efficiency. Among many possible causes, the main factor might be that PPO does 16 gradient updates (4 mini-batches and 4 update epochs) per rollout, whereas IMPALA in our setting only does 4 gradient updates. In comparison, we noticed Cleanba's IMPALA did not suffer from lower data efficiency compared to Cleanba IMPALA (Sync) architecture, meaning IMPALA can actually benefit from parallelizing actor and learner computations.

\section{Limitation}
There are several limitations to this work. First, our experiments could not completely control various other confounding settings in the reference codebase, such as optimizer settings and machine learning framework (e.g., PyTorch, JAX). For example, Cleanba's PPO and IMPALA use different learning rates indicated in their respective literature, making it difficult to compare PPO and IMPALA directly. We attempted to make a direct comparison by running Cleanba PPO with Cleanba IMPALA's setting and found it made PPO's data efficiency significantly worse -- this could suggest the IMPALA's setting is well-tuned for IMPALA but brittle to PPO (Appendix~\ref{appendix:direct-ppo-impala}). Second, our finding that parallelizing actor and learner computation hurts PPO's data efficiency is specific to the PPO's default Atari hyperparameter setting, and it could perhaps be tuned in ways in which opposite findings can be drawn. That said, the main purpose of this work is not hyperparameter tuning. Rather, it is creating a codebase that replicates prior results and makes training reproducible, efficient, and scalable across more powerful hardware.

\section{Conclusion}
This paper presents Cleanba, a new distributed deep reinforcement learning platform. Our analysis shows that Cleanba's more principled architecture can circumvent reproducibility issues in IMPALA's architecture. Our Atari experiments demonstrate that Cleanba's PPO and IMPALA accurately replicate prior work but have faster training time and are highly reproducible across different hardware settings. We believe that Cleanba will be a valuable platform for the research community to conduct future distributed RL research.

\newpage

\section*{Reproducibility Statement}
Ensuring Cleanba's results are reproducible is a central theme in our paper. To this end, we have taken several measures to improve reproducibility:

\begin{enumerate}
    \item \textbf{Open-source repository}: we made source code available at \url{https://github.com/vwxyzjn/cleanba}.
    The dependencies of the experiments are pinned, and our repository contains detailed instructions on replicating all Cleanba experiments presented in this paper.
    \item \textbf{Reproducible architecture}: as demonstrated in Section~\ref{sec:cleanba-arch}, Cleanba introduces a more principled approach to understanding distributed DRL and gives clear expectations on where the rollout data comes from, making it easier to reason about the reproducibility of distributed DRL.
    \item \textbf{Experiments on different hardware}: as demonstrated in Section~\ref{sec:experiments-atari}, we also conducted experiments showing Cleanba's PPO and IMPALA variants can obtain near-identical data efficiency on different hardware, further demonstrating that this work is highly reproducible.
\end{enumerate}

In sum, we have tried to make our work as transparent and reproducible as possible. By leveraging the source code, details provided in the main paper, and appendix, researchers should be well-equipped to reproduce or extend upon our findings.

\bibliography{iclr2024_conference}
\bibliographystyle{iclr2024_conference}
\newpage
\appendix

\section{Preliminaries}
Let us consider the RL problem in a \emph{Markov Decision Process (MDP)}~\citep{puterman2014markov}, where $\mathcal{S}$ is the state space and $\mathcal{A}$ is the action space. The agent performs some actions to the environment, and the environment transitions to another state according to its \emph{dynamics} $P (s^{\prime} \mid s, a): \mathcal{S}  \times \mathcal{A} \times \mathcal{S} \rightarrow [0, 1]$. The environment also provides a scalar reward according to the reward function $R: \mathcal{S} \times \mathcal{A} \rightarrow \mathbb{R}$, and the agent attempts to maximize the expected discounted return following a policy $\pi$: 
\begin{gather}
        J(\pi) = \mathbb{E}_{\tau}\left[G(\tau) \right]  \label{eq:rl-objective}
\\
\text { where } \tau \text { is the trajectory } \left(s_{0}, a_{0}, r_{0},  \dots, s_{T-1}, a_{T-1}, r_{T-1}, s_{T}\right) \nonumber \\
\text { and } s_{0} \sim \rho_{0}, s_t \sim P(\cdot \vert s_{t-1}, a_{t-1}), a_t \sim \pi_{\theta}(\cdot \vert s_t), r_{t}=r\left(s_{t}, a_{t}\right) \nonumber
\end{gather}

PPO~\citep{schulman2017proximal} is a popular algorithm that proposes a clipped policy gradient objective to help avoid unstable updates~\citep{schulman2017proximal,schulman2015trust}:
\begin{align}
    J^\text{CLIP}(\pi_{\theta}) =\mathbb{E}_{\tau}\left[\sum_{t=0}^{T-1}\min \left(r_t(\theta) \hat{A}_\pi^{\operatorname{adv}}(s_t, a_t), \operatorname{clip}\left(r_t(\theta), 1-\epsilon, 1+\epsilon\right) \hat{A}_\pi^{\operatorname{adv}}(s_t, a_t)\right)\right]
\end{align}

where $\pi_{\theta_{\text{old}}}$ is the policy parameter before the update,  $r_t(\theta) = \frac{\pi_{\theta}(a_t \mid s_t)}{\pi_{\theta_{\text{old}}}(a_t \mid s_t)}$, $\hat{A}_\pi^{\operatorname{adv}}$ is an advantage estimator called Generalized Advantage Estimator~\citep{schulman2015high}, and $\epsilon$ is PPO's clipped coefficient. During the optimization phase, the agent also learns the value function and maximizes the policy's entropy, therefore optimizing the following joint objective:
\begin{align}
    J^\text{JOINT}(\theta) = J^\text{CLIP}(\pi_{\theta}) - c_1J^\text{VF}(\theta) + c_2S[\pi_{\theta}], \label{eq:full-objective} 
\end{align}
where $c_1, c_2$ are coefficients, $S$ is an entropy bonus, and $J^\text{VF}$ is the squared error loss for the value function associated with $\pi_{\theta}$. Algorithm~\ref{alg:ppo} shows the pseudocode of PPO that more accurately reflects how PPO is implemented in the original codebase\footnote{\url{https://github.com/openai/baselines}}. For more detail on PPO's implementation, see \citep{shengyi2022the37implementation}. 
Given this pseudocode, the following list unifies the nomenclature/terminology of PPO's key hyperparameters.

\begin{tiny}
\begin{algorithm}[t]
\caption{Proximal Policy Optimization}\label{alg:ppo}
\begin{algorithmic}[1]
\State \textbf{Initialize} environment $E$ containing \texttt{local\_num\_envs} parallel sub-environments \label{ppo:n-envs}
\State \textbf{Initialize} policy parameters $\theta_{\pi}$, value parameters $\theta_{v}$, optimizer $O$
\State \textbf{Initialize} observation $s_{next}$, done flag $d_{next}$
\For {$i$ = 0,1,2,..., $I$}
    \State \textbf{Set} $\mathcal{D} = (s, a, \log \pi (a|s), r, d, v)$ as tuple of 2D arrays
    \For {$t$ = 0,1,2,..., \texttt{num\_steps}} \Comment{Rollout Phase} \label{ppo:nsteps}
        \State \textbf{Cache} $o_{t} = s_{next}$ and $d_{t} = d_{next}$
        \State \textbf{Get} $a_{t} \sim  \pi(\cdot| s_t; \theta_{\pi})$ and $v_t = v(s_t; \theta_{v})$\label{ppo:sample}
        \State \textbf{Step} simulator: $s_{next}, r_t, d_{next} = E.step(a_t)$\label{ppo:env-step}
        \State \textbf{Store} $s_t, d_t, v_t, a_t, \log \pi(a_t | s_t; \theta_{\pi}), r_t$ in $\mathcal{D}$\label{ppo:storage}
    \EndFor
    \State \textbf{Estimate} next value $v_{next} = v(s_{next})$ \label{ppo:bootstrap} \Comment{Learning Phase}
    \State \textbf{Compute} advantage $\hat{A_\pi^{\operatorname{adv}}}$ and return $R$ using $\mathcal{D}$ and $v_{next}$
    \State {\textbf{Prepare} the batch $\mathcal{B} = {\mathcal{D}, \hat{A_\pi^{\operatorname{adv}}}, R}$ and flatten $\mathcal{B}$}\label{ppo:batch_size}
    \For {{$epoch$ = 0,1,2,..., \texttt{update\_epochs}}} \label{ppo:epochs}
    \For {{mini-batch $\mathcal{M}$ of size $m$ in $\mathcal{B}$}} \label{ppo:mb}
            \State \textbf{Normalize} advantage $\mathcal{M}.\hat{A_\pi^{\operatorname{adv}}}$
            \State \textbf{Compute} policy loss $L^\pi$, value loss $L^V$, and entropy loss $L^S$ using $\mathcal{M}$
            \State \textbf{Back-propagate} joint loss $L = -L^\pi + c_1 L^V - c_2 L^S$
            \State \textbf{Clip} maximum gradient norm of $\theta_\pi$ and $\theta_v$ to $0.5$
            \State \textbf{Step} optimizer $O$ w.r.t. $\theta_{\pi}$ and $\theta_{v}$  \label{ppo:opt-step}
        \EndFor
    \EndFor
\EndFor
\end{algorithmic}
\end{algorithm}
\end{tiny}
\begin{itemize}
    \item \texttt{world\_size} is the number of instances of training processes; typically this is 1 (e.g., you have a single GPU).
    \item \texttt{local\_num\_envs} is the number of parallel environments PPO interacts within an instance of the training process (see line~\ref{ppo:n-envs}). \texttt{num\_envs} = \texttt{world\_size} $*$  \texttt{local\_num\_envs} is the total number of environments across all training instances.  
    \item \texttt{num\_steps} is the number of steps in which the agent samples a batch of \texttt{local\_num\_envs} actions and receives a batch of \texttt{local\_num\_envs} next observations, rewards, and done flags from the simulator (see line~\ref{ppo:nsteps}), where the done flags signal if the episodes are terminated or truncated. \texttt{num\_steps} has many names, such as the ``sampling horizon''~\citep{stooke2018accelerated} and ``unroll length''~\citep{freeman2021brax}.
    \item \texttt{local\_batch\_size} is the batch size calculated as \texttt{local\_num\_envs} $*$  \texttt{num\_steps} within an instance of the training process (\texttt{local\_batch\_size} is the size of the $\mathcal{B}$ in line~\ref{ppo:batch_size}).         \item \texttt{batch\_size} = \texttt{world\_size} $*$  \texttt{local\_batch\_size} is the aggregated batch size across all training instances. 
    \item \texttt{update\_epochs} is the number of update epochs that the agent goes through the training data in $\mathcal{B}$ (see line~\ref{ppo:epochs}).
    \item \texttt{num\_minibatches} is the number of mini-batches that PPO splits $\mathcal{B}$ into (see line~\ref{ppo:mb}).
    \item \texttt{local\_minibatch\_size} is $m = $ \texttt{local\_batch\_size} $/$ \texttt{num\_minibatches}, the size of each mini-batch $\mathcal{M}$ (see line~\ref{ppo:mb}). \texttt{minibatch\_size} = \texttt{world\_size} $*$  \texttt{local\_minibatch\_size} is the aggregated batch size across all training instances.
\end{itemize}

To make understanding more concrete, let us consider an example of Atari training. Typically, PPO uses a single training instance (i.e., \texttt{world\_size} = 1), \texttt{local\_num\_envs} = \texttt{num\_envs} = 8,  and \texttt{num\_steps} = 128. In the rollout phase (line~\ref{ppo:nsteps}-\ref{ppo:storage}), the agent collects a batch of $8*128=1024$ data points in $\mathcal{D}$. Then, suppose \texttt{num\_minibatches} = 4, $\mathcal{D}$ is evenly split to 4 mini-batches of size $m = 1024 / 4 = 256$. Next, if $K=4$,  the agent would perform $K$ * \texttt{num\_minibatches} = 16 gradient updates in the learning phase (line~\ref{ppo:bootstrap}-\ref{ppo:opt-step}).

We consider two options to scale to larger training data. \textbf{Option 1} is to increment \texttt{local\_num\_envs} -- the agent interacts with more environments, and as a result, the training data is larger. The second option is to increment \texttt{world\_size} -- have two or more copies of Algorithm~\ref{alg:ppo} running in parallel and average the gradient of the copies in line~\ref{ppo:opt-step}. \textbf{Option 2} is especially desirable when the users want to leverage more computational resources, such as GPUs.

Note that both options can be equivalent \emph{in terms of hyperparameters}. For example, when setting \texttt{world\_size} = 2, the agent effectively interacts with two distinct sets of \texttt{local\_num\_envs} environments, making its \texttt{num\_envs} doubled. To make option 1 achieve the same hyperparameters, we just need to double its \texttt{local\_num\_envs}. Below is a table summarizing the resulting hyperparameters of both options.

\begin{table}[h]
\centering
    \begin{tabular}{lp{4cm}p{4cm}}
    \toprule
    Hyperparameter & \textbf{Option 1}: Increment \texttt{local\_num\_envs} &  \textbf{Option 2}: Increment  \texttt{world\_size} \\
    \midrule
    \texttt{world\_size} & 1 & 2 \\
    \texttt{local\_num\_envs} & 120 & 60 \\
    \texttt{num\_envs} & \textbf{120} & \textbf{120}  \\
    \texttt{num\_steps} & 128 & 128  \\
    \texttt{local\_batch\_size}  & 15360 & 7680  \\
    \texttt{batch\_size} & \textbf{15360} & \textbf{15360}  \\
    \texttt{num\_minibatches} & 4 & 4  \\
    \texttt{local\_minibatch\_size} & 3840 & 1920  \\
    \texttt{minibatch\_size} & \textbf{3840} & \textbf{3840}  \\
    \bottomrule\\
    \end{tabular}
\end{table}

Importantly, we can get the same hyperparameter configuration for PPO by adjusting \texttt{local\_num\_envs} and \texttt{world\_size} accordingly. That is, we can obtain the same \texttt{num\_envs}, \texttt{batch\_size}, and \texttt{minibatch\_size} core hyperparameters.

\section{Detailed experiment settings}
\label{appendix:detailed-setting}
For the experiments, the PPO and IMPALA's hyperparameters can be found in Table~\ref{tab:cleanrl-ppo-atari-params}. The Vtrace implementation can be found in \texttt{rlax}\footnote{\url{https://github.com/deepmind/rlax/blob/b53c6510c8b2cad6b106b6166e22aba61a77ee2f/rlax/_src/vtrace.py\#L162-L193}}.

\begin{table}[ht]
\centering
\caption{PPO hyperparameters.}
\begin{tabular}{ll} 
\toprule
Parameter Names  & Parameter Values\\
\midrule
$N_\text{total}$ Total Time Steps & 50,000,000  \\ 
$\alpha$ Learning Rate &  \colorbox{LightAquamarine}{0.00025 Linearly Decreased to 0} \\
$N_\text{envs}$ Number of Environments & 128 \\
$N_\text{steps}$ Number of Steps per Environment & \colorbox{LightSalmon}{128}  \\
$\gamma$ (Discount Factor) & 0.99 \\ 
$\lambda$ (for GAE) & 0.95 \\ 
$N_\text{mb}$ Number of Mini-batches & 4 \\
$K$ (Number of PPO Update Iteration Per Epoch)& 4 \\
$\varepsilon$ (PPO's Clipping Coefficient) & 0.1 \\
$c_1$ (Value Function Coefficient)& 0.5\\
$c_2$ (Entropy Coefficient)& 0.01\\
$\omega$ (Gradient Norm Threshold)& \colorbox{LightGoldenrod}{0.5} \\
Value Function Loss Clipping
& False\\
Optimizer Setting & \colorbox{LightOrchid}{Adam optimizer with $\epsilon=0.00001$}\\
\bottomrule
\end{tabular}
\label{tab:cleanrl-ppo-atari-params}
\caption{IMPALA hyperparameters.}
\begin{tabular}{ll} 
\toprule
Parameter Names  & Parameter Values\\
\midrule
$N_\text{total}$ Total Time Steps & 50,000,000  \\ 
$\alpha$ Learning Rate &  \colorbox{LightAquamarine}{0.0006 Linearly Decreased to 0} \\
$N_\text{envs}$ Number of Environments & 128 \\
$N_\text{steps}$ Number of Steps per Environment & \colorbox{LightSalmon}{20}  \\
$\gamma$ (Discount Factor) & 0.99 \\ 
$\lambda$ (mixing parameter) & 1.0 \\ 
$N_\text{mb}$ Number of Mini-batches & 4 \\
$\rho$ (Clip Threshold for Importance Ratios) & 1.0 \\
$\rho_{pg}$ (Clip Threshold for Policy Gradient Importance Ratios) & 1.0 \\
$c_1$ (Value Function Coefficient)& 0.5\\
$c_2$ (Entropy Coefficient)& 0.01\\
$\omega$ (Gradient Norm Threshold)& \colorbox{LightGoldenrod}{40.0} \\
Optimizer Setting & \colorbox{LightOrchid}{RMSprop optimizer with $\epsilon=0.01$,}\\
& \colorbox{LightOrchid}{$\text{decay}=0.99$}\\
\bottomrule
\end{tabular}
\label{tab:cleanrl-IMPALA-atari-params}
\end{table}

\section{\texttt{moolib} Experiments}
\label{appendix:moolib_twosets}
By default, \texttt{moolib} uses 256 environments, 10 actor CPUs, and a single GPU. We followed the recommended scaling instructions to add 8 training GPU-powered peers, which in total used 2048 environments, 80 actor CPUs, and 8 GPUs. While the training time was reduced to about 27 minutes, sample efficiency dropped, and it obtained a catastrophic 28.51\% median HNS after 200M frames. We suspected the drop was due to the 2048 environments used, so we set the total number of environments back to 256. Furthermore, we did not restrict \texttt{moolib} to use 50 CPUs because we worried it might change the learning behaviors due to the issues mentioned in Section~\ref{sec:impala-reproducibility-issues}, so we kept the default scaling to 80 CPUs. For comparison with \texttt{moolib}, \texttt{monobeast} experiments also use 80 CPUs.

\begin{figure}[t]
    \centering
    \includegraphics[width=0.90\linewidth]{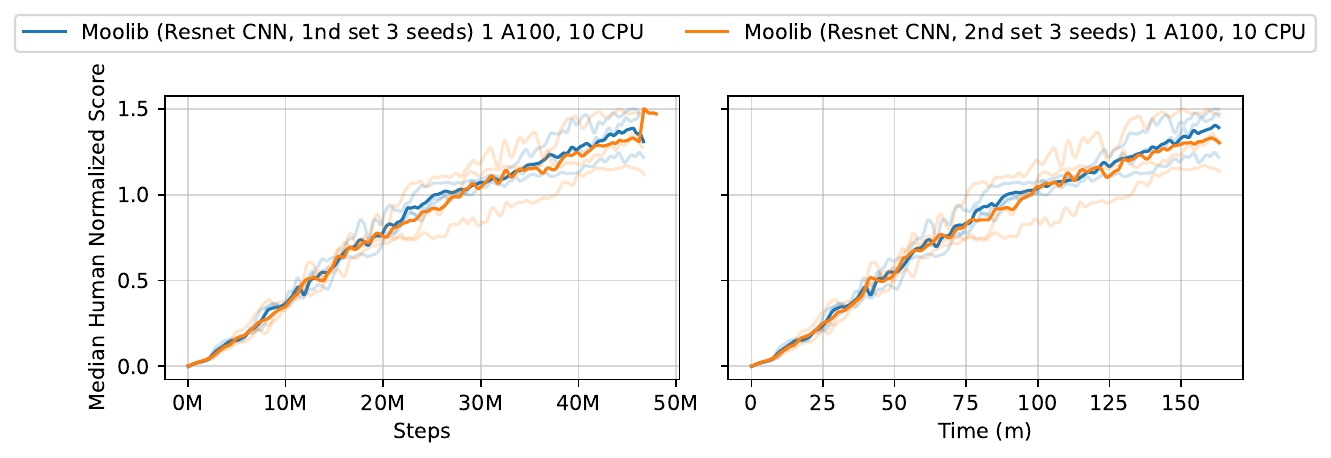}
    \includegraphics[width=0.90\linewidth]{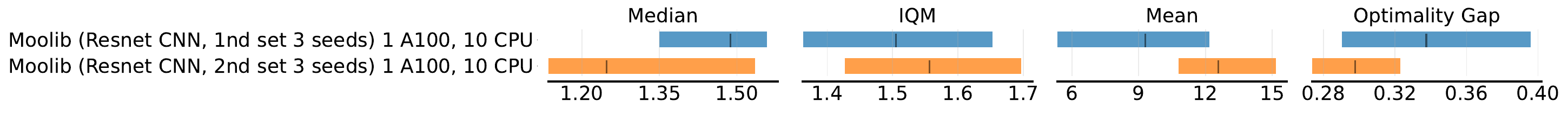}
    \includegraphics[width=0.99\linewidth]{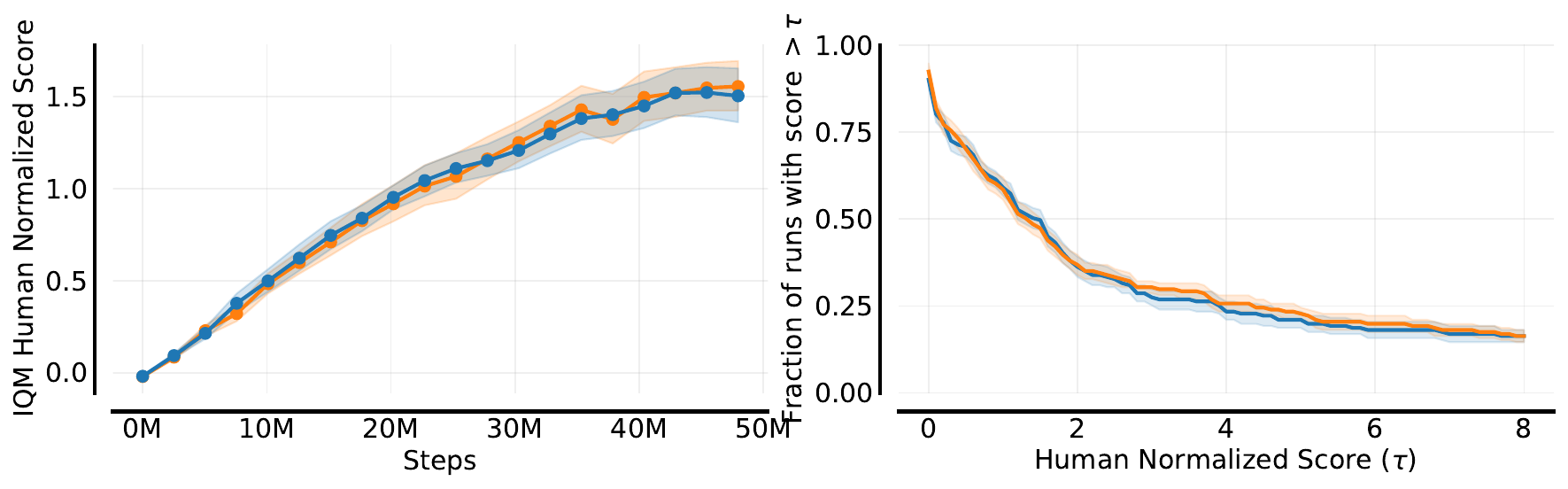}
    \caption{Top figure: the median human-normalized scores of the two sets of \texttt{moolib} experiments. Middle figure: the IQM human-normalized scores and performance profile~\citep{agarwal2021deep}. Bottom figure: the average runtime in minutes and aggregate human normalized score metrics with 95\% stratified bootstrap CIs.}
    \label{fig:moolib-two-set-experiments}
\end{figure}

We conducted two sets of \texttt{moolib} experiments and reported the set with a lower median and higher IQM, as shown in Figure~\ref{fig:moolib-two-set-experiments} for legacy reasons. During our debugging, we found the Asteroids experiments in the first set of \texttt{moolib} experiments to obtain high scores, but we ran Asteroids specifically for ten random seeds and found lower scores; this suggests the Asteroids experiments in the first set were likely due to lucky random seeds, so we re-run the \texttt{moolib} experiments.

\section{The effect of different wrappers on \texttt{moolib}'s performance}
\label{appendix:wrapper-matter}
\begin{figure}[t]
    \includegraphics[width=\textwidth]{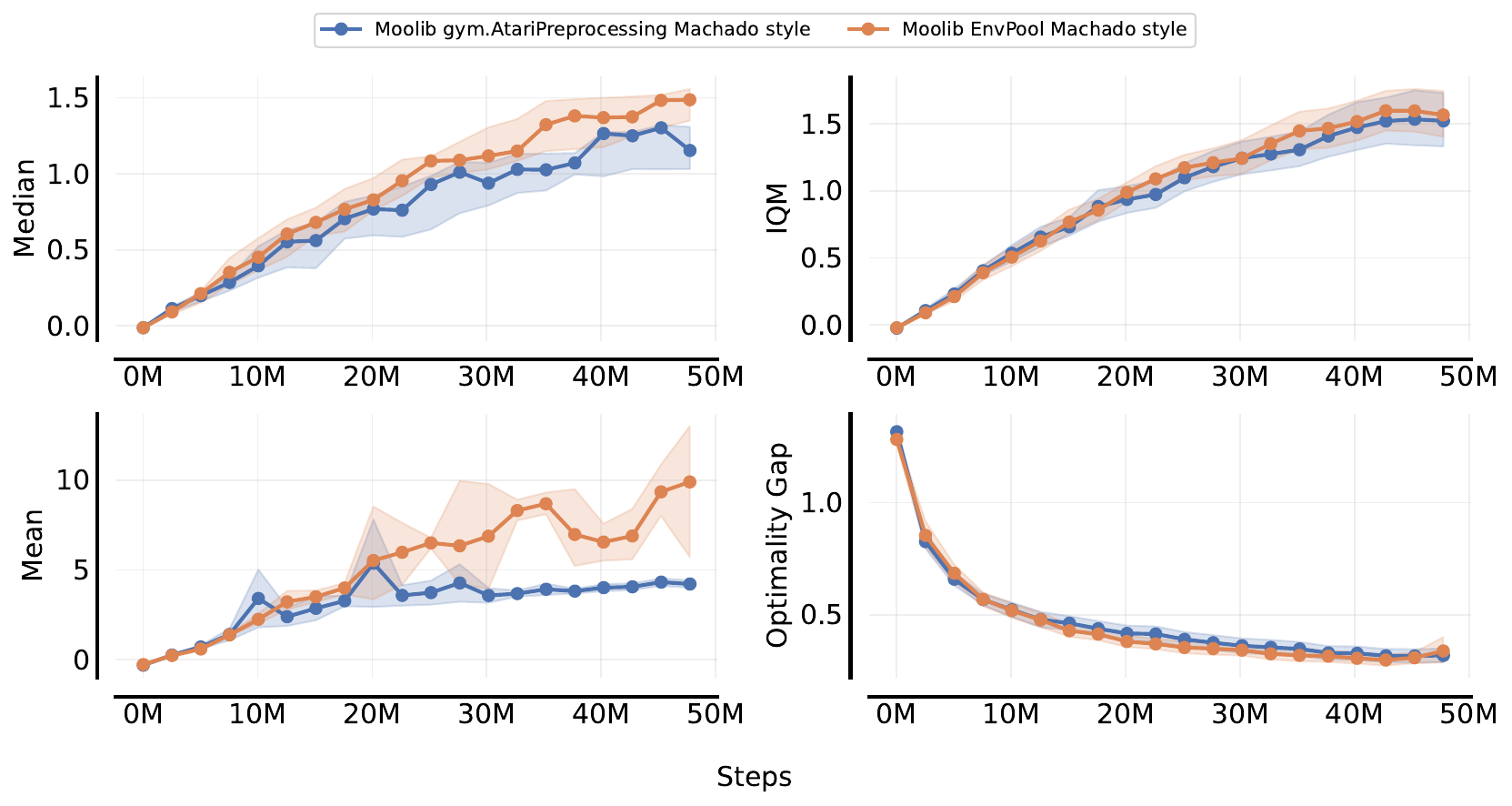}
    \includegraphics[width=0.99\linewidth]{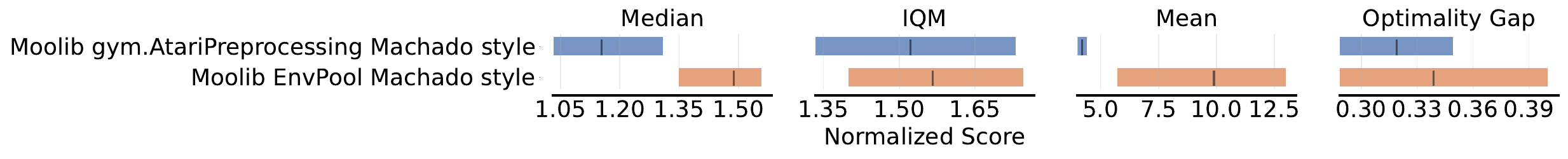}
    \caption{Atari wrappers matter. When using  \texttt{gym.AtariPreprocessing} wrappers with a comparable setting to our EnvPool setup, we found \texttt{moolib}to have lower median and mean HNS, although IQM is roughly the same. }
    \label{fig:different-wrappers}
\end{figure}

The Atari wrappers can be important to the agent's performance. As a preliminary study, we used \texttt{moolib}'s default Atari wrappers\footnote{\href{https://github.com/facebookresearch/moolib/blob/06e7a3e80c9f52729b4a6159f3fb4fc78986c98e/examples/atari/atari\_preprocessing.py}{\url{https://github.com/facebookresearch/moolib/blob/main/examples/atari/atari\_preprocessing.py}}} implemented with \texttt{gym.AtariPreprocessing} to run experiments and compare the results with the ones presented in the main text of the paper. As shown in Figure~\ref{fig:different-wrappers}, Atari wrappers matter -- \texttt{moolib}'s default \texttt{AtariPreprocessing} wrappers result in lower median and mean HNS, although IQM is roughly the same. To make a fair comparison, the experiments presented in the main text all use the same \envpool~Atari wrappers.

\section{Direct PPO and IMPALA comparison}
\begin{figure}[t]
    \includegraphics[width=\textwidth]{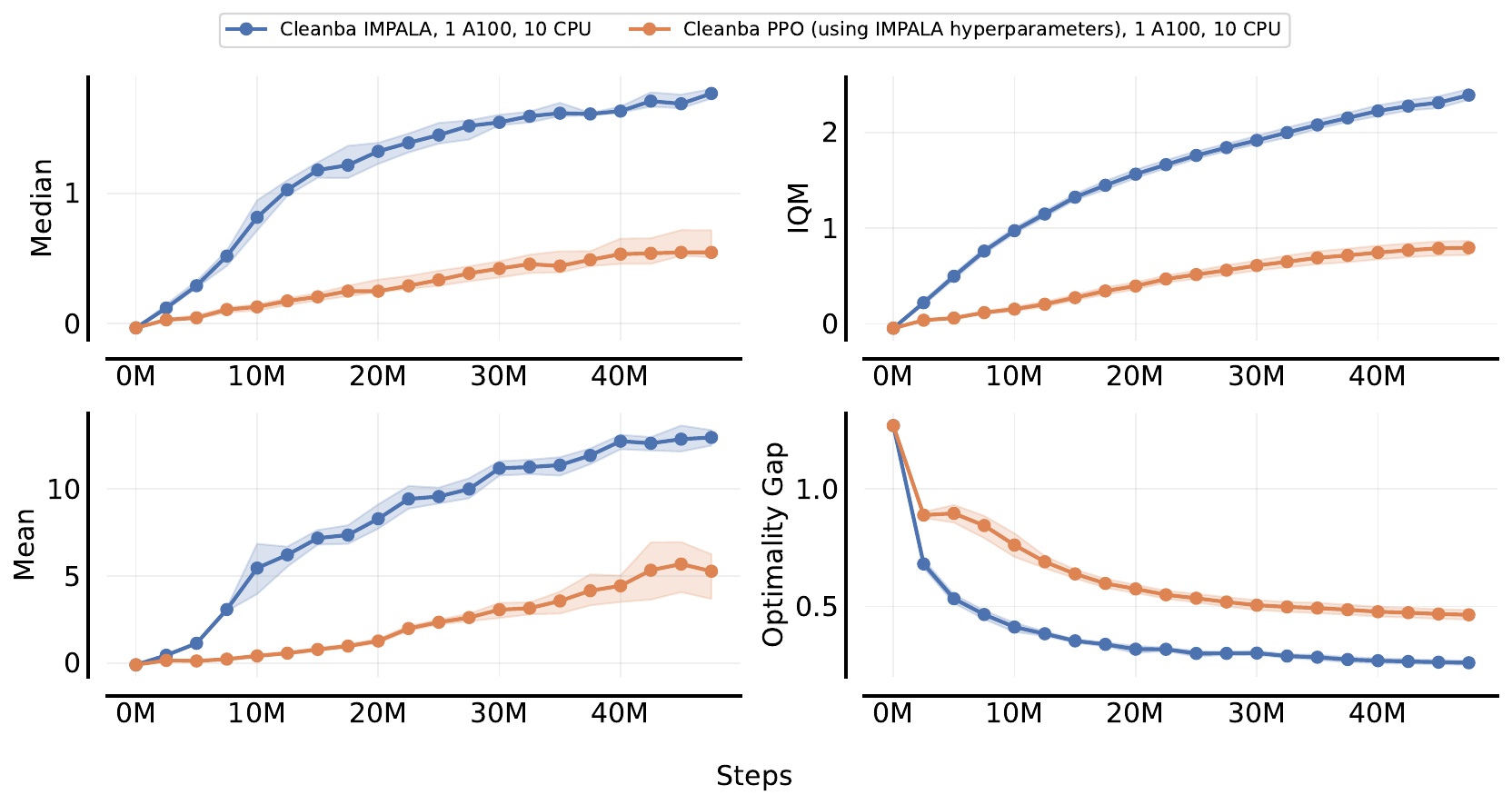}
    \includegraphics[width=0.99\linewidth]{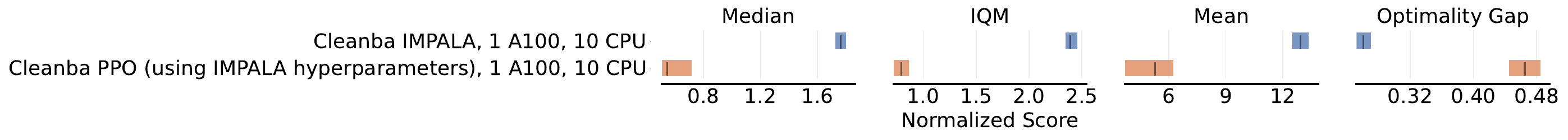}
    \caption{A direct PPO and IMPALA comparison. Running Cleanba PPO using Cleanba IMPALA's setting. Note that this is not a fair comparison because Cleanba IMPALA's setting is likely well-tuned IMPALA setting but not well-tuned PPO setting}
    \label{fig:ppo-impala-direct}
\end{figure}

To make a direct (but not fair) comparison between PPO and IMPALA, we ran Cleanba PPO using IMPALA's settings and the results can be found at Figure~\ref{fig:ppo-impala-direct}.
\label{appendix:direct-ppo-impala}






\pagebreak

\section{Large Batch Size Training} Cleanba can also scale to the hundreds of GPUs in multi-host and multi-process environments by leveraging the \texttt{jax.distributed} package, allowing us to explore training with even larger batch sizes. Here we use an earlier version of the codebase to conduct experiments with 16, 32, 64, and 128 A100 GPUs. For convenience, we also adjust a few settings: 1) turn off the learning rate annealing, 2) run for 100M steps instead of the standard 50M steps, and 3) keep doubling the \texttt{num\_envs}, \texttt{batch\_size}, and \texttt{minibatch\_size} with a larger number of GPUs.

\begin{figure}[t]
    \centering
    \includegraphics[width=1\linewidth]{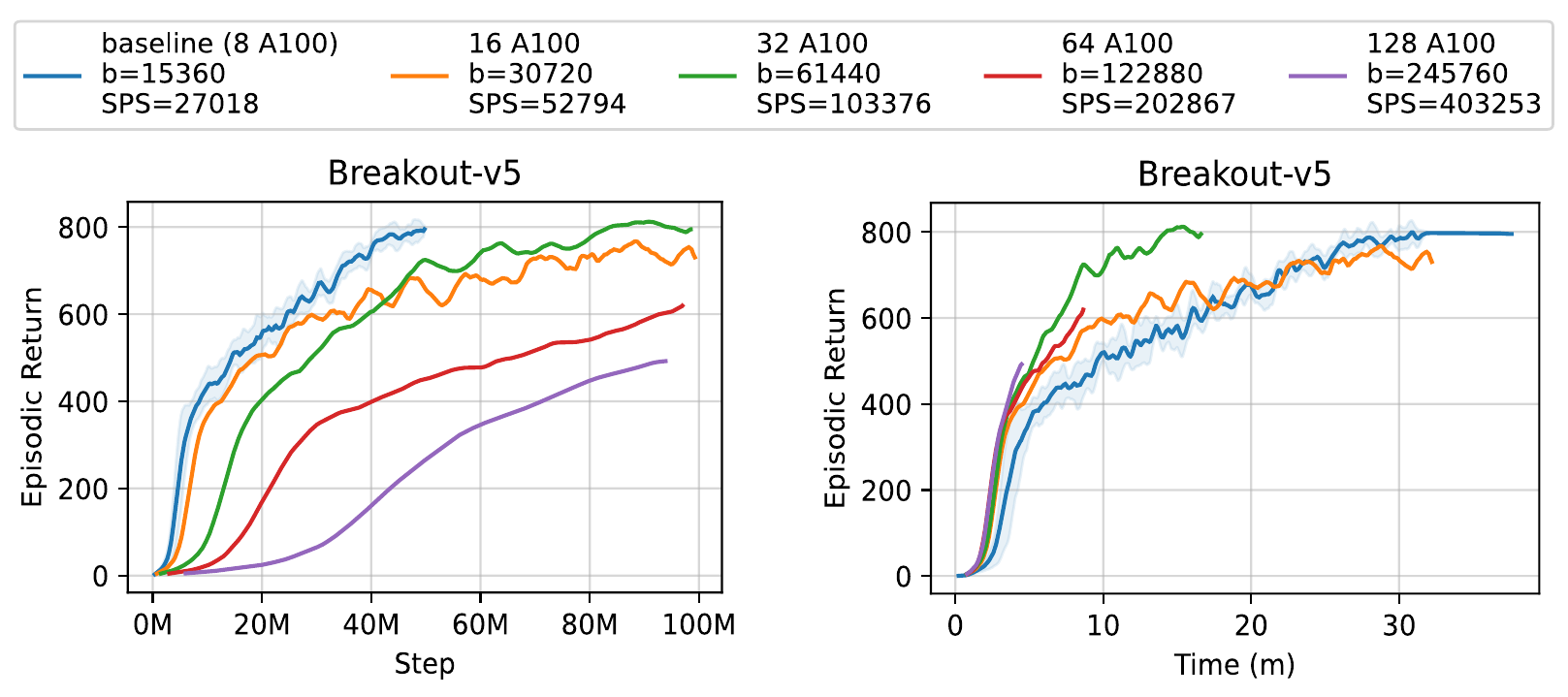}
    \caption{Cleanba's results from large batch size training.  \texttt{b=15360} denotes \texttt{batch\_size=15360}. 
    }
    \label{fig:largeb}
\end{figure}

Due to hardware scheduling constraints, we only ran the experiments for 1 random seed. The results are shown in Figure~\ref{fig:largeb}. We make the following observations:
\begin{itemize}
    \item \textbf{Linear scaling w/ 93\% of ideal scaling efficiency.} As we increased the number of GPUs to 16, 32, 64, 128, we observed a linear scaling in steps per second (SPS) in Cleanba achieving 93\% of the ideal scaling efficiency. This is likely empowered by the fast connectivity offered by NVIDIA GPUDirect RDMA (remote direct memory access) in Stability AI's HPC. When using 128 GPUs, the agent has an SPS of 403253, translating to over \emph{1.6M FPS} in Breakout.
    \item \textbf{Small batch sizes train more efficiently.} As we increase batch sizes, particularly in the first 40M steps, the sample efficiency tends to decline. This outcome is unsurprising, given that the initial policy is random and Breakout initially has limited explorable game states. In this case, the data in the batch is going to have less diverse data, which makes the large batch size less valuable.  
    \item \textbf{Large batch sizes train more quickly.} Like \citep{mccandlish2018empirical}, we find increasing the batch size does make the agent reach some given scores faster. This suggests that we could always increase the batch size to obtain shorter training times if sample efficiency is not a concern.
\end{itemize}
While we observed limited benefits of scaling Cleanba to use 128 GPUs, the objective of the scaling experiments is to show we can scale to large batch sizes. Given a more challenging task, the training data is likely going to be more diverse and have a higher \emph{gradient noise scale}~\citep{mccandlish2018empirical}, which would help the agent utilize large batch sizes more efficiently, resulting in a reduced decline in sample efficiency.

\begin{figure}[t]
    \centering
    \includegraphics[width=0.5\linewidth]{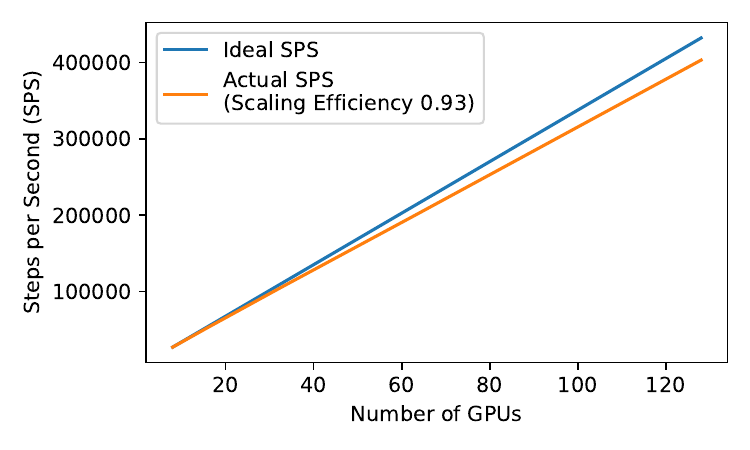}
    \caption{Cleanba's SPS scaling results from large batch size training. }
    \label{fig:largeb-scaling}
\end{figure}

\section{torchbeast logs}
\label{sec:torchbeast_logs}

\begin{minted}[frame=single,framesep=10pt,fontsize=\mysmall]{python}
$ python -m torchbeast.monobeast_study \
--num_actors 80 \
--total_steps 10000000 \
--learning_rate 0.0006 \
--epsilon 0.01 \
--entropy_cost 0.01 \
--batch_size 8 \
--unroll_length 240 \
--num_threads 1 \
--env Pong-v5
actor_index 32 initial policy_version 8 policy_version after rollout 20
actor_index 13 initial policy_version 8 policy_version after rollout 20
actor_index 57 initial policy_version 8 policy_version after rollout 20
actor_index 12 initial policy_version 8 policy_version after rollout 21
actor_index 51 initial policy_version 8 policy_version after rollout 21
actor_index 2 initial policy_version 8 policy_version after rollout 21
actor_index 56 initial policy_version 8 policy_version after rollout 21
actor_index 38 initial policy_version 9 policy_version after rollout 21
actor_index 37 initial policy_version 9 policy_version after rollout 22
actor_index 59 initial policy_version 9 policy_version after rollout 22
actor_index 9 initial policy_version 9 policy_version after rollout 22
actor_index 69 initial policy_version 9 policy_version after rollout 22
actor_index 35 initial policy_version 9 policy_version after rollout 22
actor_index 66 initial policy_version 9 policy_version after rollout 22
actor_index 10 initial policy_version 9 policy_version after rollout 22
actor_index 55 initial policy_version 10 policy_version after rollout 22
actor_index 53 initial policy_version 10 policy_version after rollout 22
actor_index 46 initial policy_version 10 policy_version after rollout 22
actor_index 54 initial policy_version 10 policy_version after rollout 23
actor_index 50 initial policy_version 10 policy_version after rollout 23
actor_index 8 initial policy_version 10 policy_version after rollout 23
actor_index 64 initial policy_version 10 policy_version after rollout 23
actor_index 77 initial policy_version 10 policy_version after rollout 23
actor_index 3 initial policy_version 11 policy_version after rollout 23
actor_index 7 initial policy_version 11 policy_version after rollout 23
actor_index 28 initial policy_version 11 policy_version after rollout 23
actor_index 49 initial policy_version 11 policy_version after rollout 23
actor_index 16 initial policy_version 11 policy_version after rollout 23
actor_index 24 initial policy_version 11 policy_version after rollout 23
actor_index 11 initial policy_version 11 policy_version after rollout 23
actor_index 14 initial policy_version 11 policy_version after rollout 23
actor_index 43 initial policy_version 13 policy_version after rollout 26
actor_index 58 initial policy_version 13 policy_version after rollout 26
actor_index 23 initial policy_version 13 policy_version after rollout 26
actor_index 29 initial policy_version 13 policy_version after rollout 26
actor_index 68 initial policy_version 13 policy_version after rollout 26
actor_index 75 initial policy_version 14 policy_version after rollout 26
actor_index 48 initial policy_version 14 policy_version after rollout 27
actor_index 67 initial policy_version 14 policy_version after rollout 27
actor_index 5 initial policy_version 14 policy_version after rollout 27
actor_index 18 initial policy_version 14 policy_version after rollout 27
actor_index 41 initial policy_version 15 policy_version after rollout 27
actor_index 78 initial policy_version 14 policy_version after rollout 27
actor_index 15 initial policy_version 15 policy_version after rollout 27
actor_index 34 initial policy_version 15 policy_version after rollout 27
actor_index 45 initial policy_version 15 policy_version after rollout 28
actor_index 22 initial policy_version 15 policy_version after rollout 28
actor_index 4 initial policy_version 16 policy_version after rollout 28
actor_index 6 initial policy_version 16 policy_version after rollout 28
actor_index 20 initial policy_version 16 policy_version after rollout 28
actor_index 39 initial policy_version 16 policy_version after rollout 28
actor_index 33 initial policy_version 16 policy_version after rollout 29
actor_index 74 initial policy_version 16 policy_version after rollout 29
actor_index 60 initial policy_version 16 policy_version after rollout 29
actor_index 42 initial policy_version 17 policy_version after rollout 29
actor_index 72 initial policy_version 17 policy_version after rollout 30
actor_index 25 initial policy_version 17 policy_version after rollout 30
actor_index 31 initial policy_version 17 policy_version after rollout 30
actor_index 19 initial policy_version 17 policy_version after rollout 30
actor_index 1 initial policy_version 18 policy_version after rollout 31
actor_index 79 initial policy_version 18 policy_version after rollout 31
actor_index 65 initial policy_version 18 policy_version after rollout 31
actor_index 73 initial policy_version 18 policy_version after rollout 31
actor_index 36 initial policy_version 18 policy_version after rollout 31
actor_index 21 initial policy_version 18 policy_version after rollout 31
actor_index 0 initial policy_version 18 policy_version after rollout 31
actor_index 30 initial policy_version 18 policy_version after rollout 31
actor_index 44 initial policy_version 18 policy_version after rollout 31
actor_index 63 initial policy_version 19 policy_version after rollout 31
actor_index 76 initial policy_version 19 policy_version after rollout 32
actor_index 47 initial policy_version 19 policy_version after rollout 32
actor_index 52 initial policy_version 19 policy_version after rollout 32
actor_index 26 initial policy_version 19 policy_version after rollout 32
actor_index 71 initial policy_version 19 policy_version after rollout 32
actor_index 70 initial policy_version 19 policy_version after rollout 32
actor_index 17 initial policy_version 20 policy_version after rollout 32
actor_index 62 initial policy_version 20 policy_version after rollout 33
actor_index 40 initial policy_version 20 policy_version after rollout 33
actor_index 27 initial policy_version 20 policy_version after rollout 33
actor_index 13 initial policy_version 20 policy_version after rollout 33
actor_index 57 initial policy_version 20 policy_version after rollout 33
actor_index 32 initial policy_version 20 policy_version after rollout 33
actor_index 51 initial policy_version 21 policy_version after rollout 33
actor_index 61 initial policy_version 20 policy_version after rollout 33
actor_index 2 initial policy_version 21 policy_version after rollout 33
actor_index 56 initial policy_version 21 policy_version after rollout 34
actor_index 12 initial policy_version 21 policy_version after rollout 34


$ python -m torchbeast.monobeast_study \
--num_actors 80 \
--total_steps 10000000 \
--learning_rate 0.0006 \
--epsilon 0.01 \
--entropy_cost 0.01 \
--batch_size 8 \
--unroll_length 240 \
--num_threads 1 \
--env Pong-v5  \
--learner_delay_seconds 1.0

actor_index 72 initial policy_version 9 policy_version after rollout 10
actor_index 22 initial policy_version 9 policy_version after rollout 10
actor_index 37 initial policy_version 9 policy_version after rollout 10
actor_index 41 initial policy_version 9 policy_version after rollout 10
actor_index 16 initial policy_version 9 policy_version after rollout 10
actor_index 61 initial policy_version 10 policy_version after rollout 11
actor_index 18 initial policy_version 10 policy_version after rollout 11
actor_index 13 initial policy_version 10 policy_version after rollout 11
actor_index 56 initial policy_version 10 policy_version after rollout 11
actor_index 28 initial policy_version 10 policy_version after rollout 11
actor_index 4 initial policy_version 10 policy_version after rollout 11
actor_index 7 initial policy_version 10 policy_version after rollout 11
actor_index 65 initial policy_version 10 policy_version after rollout 11
actor_index 12 initial policy_version 11 policy_version after rollout 12
actor_index 14 initial policy_version 11 policy_version after rollout 12
actor_index 5 initial policy_version 11 policy_version after rollout 12
actor_index 3 initial policy_version 11 policy_version after rollout 12
actor_index 35 initial policy_version 11 policy_version after rollout 12
actor_index 51 initial policy_version 11 policy_version after rollout 12
actor_index 0 initial policy_version 11 policy_version after rollout 12
actor_index 6 initial policy_version 11 policy_version after rollout 12
actor_index 60 initial policy_version 12 policy_version after rollout 13
actor_index 77 initial policy_version 12 policy_version after rollout 13
actor_index 48 initial policy_version 12 policy_version after rollout 13

$ python -m torchbeast.monobeast_study \
--num_actors 40 \
--total_steps 10000000 \
--learning_rate 0.0006 \
--epsilon 0.01 \
--entropy_cost 0.01 \
--batch_size 8 \
--unroll_length 240 \
--num_threads 1 \
--env Pong-v5

actor_index 34 initial policy_version 12 policy_version after rollout 18
actor_index 25 initial policy_version 13 policy_version after rollout 18
actor_index 4 initial policy_version 13 policy_version after rollout 18
actor_index 5 initial policy_version 13 policy_version after rollout 18
actor_index 14 initial policy_version 13 policy_version after rollout 18
actor_index 16 initial policy_version 13 policy_version after rollout 18
actor_index 12 initial policy_version 13 policy_version after rollout 18
actor_index 39 initial policy_version 13 policy_version after rollout 18
actor_index 30 initial policy_version 13 policy_version after rollout 18
actor_index 18 initial policy_version 13 policy_version after rollout 18
actor_index 13 initial policy_version 13 policy_version after rollout 18
actor_index 23 initial policy_version 13 policy_version after rollout 19
actor_index 35 initial policy_version 13 policy_version after rollout 19
actor_index 3 initial policy_version 14 policy_version after rollout 19
actor_index 17 initial policy_version 14 policy_version after rollout 19
actor_index 9 initial policy_version 14 policy_version after rollout 19
actor_index 6 initial policy_version 14 policy_version after rollout 19
\end{minted}

\end{document}